\documentclass{elsarticle}
\usepackage{lipsum}
\makeatletter
\def\ps@pprintTitle{%
 \let\@oddhead\@empty
 \let\@evenhead\@empty
 \def\@oddfoot{}%
 \let\@evenfoot\@oddfoot}
\makeatother

\usepackage{graphicx}
\usepackage{caption}
\usepackage{subcaption}
\usepackage{listings}
\usepackage{amssymb}
\usepackage{amsmath}

\usepackage{natbib}
\begin{document}

\begin{frontmatter}


\title{Deep CNN Frame Interpolation with Lessons Learned from Natural Language Processing}

\author{Kian Ghodoussi$^{[1]}$, Nihar Sheth$^{[2]}$, Zane Durante$^{[3]}$, Markie Wagner$^{[3]}$}

\address{University of Southern California}

\begin{abstract}
A major area of growth within deep learning has been the study and implementation of Convolutional Neural Networks. The general explanation within the deep learning community of the robustness of convolutional neural networks within image recognition rests upon the idea that CNNs are able to extract localized features. However, recent developments in fields such as Natural Language Processing are demonstrating that this paradigm may be incorrect. In this paper, we analyze the current state of the field concerning CNN's and present a hypothesis that provides a novel explanation for the robustness of CNN models. From there, we demonstrate the effectiveness of our approach by presenting novel deep CNN frame interpolation architecture that is comparable to the state of the art interpolation models with a fraction of the complexity. 
\end{abstract}


\end{frontmatter}


\section{Introduction}
\label{S:1}

In recent years, a focal point of research within the deep learning community has been the internal mechanisms of Convolutional Neural Networks (CNNs). The primary use-case of CNNs has been within the domain of image recognition; however, CNNs have found their way within a variety domains, ranging from natural language processing to deep encoding. A area of interest that seems to hold considerable potential is the internal processing and features within the CNN, i.e. the mysterious intermediate layers.

CNNs are a class of neural network that specializes in localized detection \cite{cnn-basic}. They leverage a sliding filter in order to perturb, or 'convolve' data in a theoretically beneficial way. In signal processing and classic image recognition, these filters were often hand derived, with various kernels built for edge detection at various angles. The major development CNNs introduced was the ability to use gradient descent in order to learn filter weights, as opposed to manually tuning them. This gradient-based method has proven to be extremely powerful for image recognition, enabling the detection of localized features, regardless of their positions, and has inspired a growing body of work on the the indirect applications of these filter weights. 

\citeauthor{style-transfer} demonstrated that earlier convolution layers actually detect more low level features such as texture and color pallet, while deeper layers pick up more high level features such as content. \cite{style-transfer} demonstrated that these various levels of inference can be leveraged to perform a "Style Transfer," generating a new image that minimizes the difference between a "content image's" high level features and a "style images" low level features.

\cite{transfer-learning} demonstrates that CNNs trained on image classification tasks can also serve as a powerful method of dimensionality reduction. Methods such as transfer learning, which consists of taking a pre-trained deep CNN network, dropping the final fully connected layers, and retraining a new set of fully connected layers on CNN encodings, have become a staple within the computer vision community \cite{transfer-learning}. Since then, a number of applications for these deep CNN encodings have broken out, with use cases ranging from image clustering \cite{img-clustering} to dynamic captioning \cite{img-captioning} to our use-case of frame interpolation.

Classically, frame interpolation is a method of enhancing the smoothness and potentially quality of a choppy video. However, a new use case that is slowly becoming viable is within the field of live streaming. With a robust enough model, it could be potentially possible to exponentially reduce the number of frames sent through during a live streaming video services. And where money is, machine learning goes. In Section \ref{overview}, we discuss the current work done in the field of frame interpolation. In Section \ref{nlp_sec}, we analyze recent findings that challenge the way CNN's are viewed and present a potentially new paradigm when designing Deep CNN networks. In the Section \ref{model_arch_sec}, we discuss the specifics of our model, and in section \ref{results}, we present our results.

\section{Overview of Current Frame Interpolation Models}
\label{overview}

\begin{figure}[]
\includegraphics[scale=0.25]{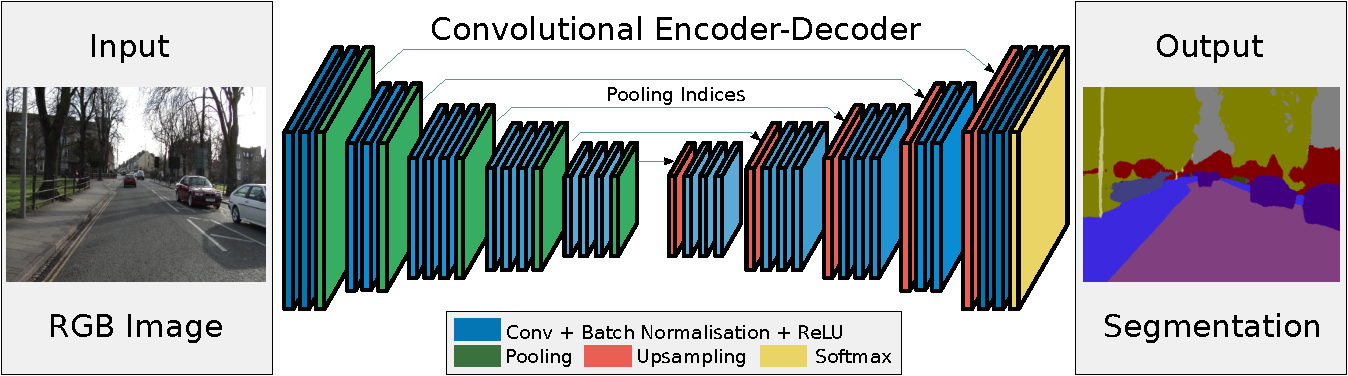}
\centering{}
\caption{Encoder Decoder Network depicted in \citeauthor{encoder_decoder_pic}}
\label{encoder-decoder}
\end{figure}

The usage of convolutional neural networks in the domain of frame interpolation is not novel proposal. \citeauthor{Niklaus_encoder} proposes the use of a classic encoder/decoder CNN architecture, similar to the architecture depicted in Figure \ref{encoder-decoder}. In this model architecture, the encoder component of the network learns a dense, low dimensional feature representation of the original frames. The low dimensional encodings of the two frames are then stacked on top of each other. The decoder network then performs up-sampling in order to predict the intermediate frame, given the two previous encodings. \cite{Niklaus_encoder} demonstrates that this method outperforms classic frame interpolation methods.

The hypothesis driving this architecture is the idea that a CNN can detect local features and represent these features in a low dimensional space. And this encoder-decoder architecture, while displaying small differences, is the most widely used approach for frame interpolation models. \citeauthor{nvidia_encoder_decoder} takes an extremely similar approach to \cite{Niklaus_encoder}, simply substituting in the LeakyRelu activation function instead of Relu (a choice we found improves performance in our model). The majority of these models define their loss functions as the mean-squared error of the predicted pixed values, i.e.: 

\begin{equation}
	Loss = \frac{1}{n} \sum_{i=0}^{n}{ (yp_{i}-y_{ij})^2 } 
    \label{mse-pixels}
\end{equation}

\citeauthor{gongvideo_autoencoder} take the encoder/decoder architecture one step further, substituting in a pretrained autoencoder, rather than training an encoder themselves. \cite{gongvideo_autoencoder} further leverages the CNN's feature detection capabilities, by replacing the mean-squared-error loss function in Equation \ref{mse-pixels} with the loss function described in Equation \ref{mse-encoding}, where ij is the $i^{th}$ pixel at the $j^{th}$ encoding layer.

\begin{equation}
	Loss = \frac{1}{n*m} \sum_{i=0, j=0}^{n, m}{ (   Encoder( yp )_{ij} - Encoder(y )_{ij} )^2 } 
    \label{mse-encoding}
\end{equation}

These models each perform exceptionally well. It is difficult to compare the various methods due to their different metrics and their focus on qualitative visual measures. By visual inspection, it is clear that each of them perform far better than the current methods of frame interpolation.

However, given recent revelations concerning the inner workings of CNN's, it appears to us that the encoder/decoder architecture may not be the optimal approach for this problem space. In the next section, we analyze trends in  the field natural language processing that inspire our model architecture. 

\section{Lessons Learned from Natural Language Processing}
\label{nlp_sec}
The key idea behind the majority of CNN architectures is the idea of CNN's detecting local features. ML practitioners add more layers and more sliding windows with the hopes of detecting more complex, more nuanced, or simply \textit{more} features. 

However, the findings in the NLP paper by \cite{Kim2014} demonstrate that CNN's may not be learning features at all. In \cite{Kim2014}, a CNN was tasked with document classification, using two different word representations. The first was using pre-trained word embeddings, which is defined by \cite{Mikolov2013a} as a method to meaningfully represent words in an M-dimensional vector space. The second representation was simply to map each word to a random M-dimensional vector. From there, the document is treated as an NxM dimensional image, where N is the number of words in the document and  M is the dimensionality of the embeddings. From there, the CNN filter simply slides down the sequence of word embeddings, as seen below. 

\begin{equation*}
N-Words
\underbrace{\left[
	\begin{array}{ccccccccccccc} 
    \\
    \left[\begin{array}{ccccccccccccc}  CNN Filter \end{array}\right]
    \\ \\ \\ \\ \\ \\ \\ \\ 
    \end{array} \right ]}_{M-Dim-Embedding}
    \label{kmaxCNN}
\end{equation*}

The hypothesis was that the CNN should be able to pick up on semantic trends within the word embeddings. As a result, the word-embedding-based CNN was expected to outperform or, at the very least, train faster than the random embedding-based model.

However, in \cite{Kim2014}, the performance and the learning curves of the two CNNs were statistically equivalent. These findings call into question the idea that CNNs are actually learning local feature representations and our interpretation of image data entirely.

We propose that images are effectively sparse data in the sense that, there are very few pixels that genuinely contribute to the meaning of an image. This proposal is a natural extension of \cite{one_pixel_attack}, which demonstrates that the alteration only 3 targeted pixels in an image changes 40\% of a Deep CNN's class predictions. 

Shifting to the sparse perspective pulls us towards the domain of natural language processing, a field that consistently deals with sparse feature sets due to the bursty nature of word usage. However, contrary to classic machine learning heuristics, it turns out this dimensionality is not necessarily a curse. One of the most popular methods in NLP is a Bag-of-Words representation, where each word is represented as a one hot vector that is the length of an arbitrary vocabulary. Bag of Words methods often lead to extremely high dimensional data relative to the number training examples: yet, when combined with a simple linear classifier such as logistic regression, outperforms many of the most complex deep learning based document classification methods. 

As a result of these developments, we propose that robustness of CNN's comes not from their ability to detect local features, but rather from their ability to 'sparsify' and represent images within a high dimensional space. In Section \ref{model_arch_sec}, we present a novel deep CNN frame interpolation that leverages these lessons learned from natural language processing.

\section{Frame Interpolation Model Architecture}
\label{model_arch_sec}

The current encoder/decoder architectures represent a significant improvement over classical methods. However, these models are predicated upon hypothesis that is treated as fact: that the robustness of a CNN is due to its detect local features. 

We build a model that challenges that hypothesis. We instead build our model around the idea that images are effectively sparse data, and CNNs are a tool to "sparsify" the frame by extrapolating pixels into a high dimensional space (Section \ref{nlp_sec}). Rather than predicting the next frame based on the features present in a deep encoding, our goal is to instead learn a high dimensional representation of images that enables the interpolation of two frames. 

\begin{figure}[]
\includegraphics[scale=0.8]{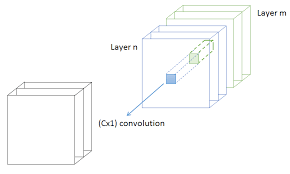}
\centering{}
\caption{Visualization of a Convolution within a Convolution Layer}
\label{conv_in_conv}
\end{figure}

Our model's first layer utilizes a convolution with a 1x1 filter. The goal of this layer is to represent RGB pixels in a higher dimensional space, before detecting any relationships between pixels. This layer is visualized in Figure \ref{conv_in_conv}.

From there, our model consists of the sequence Filter+leakyReLU+Dropout. We start with a 7x7 filter, and as we move deeper in the network, we increase the number of filters while decreasing the filter size. In the last layer, we utilize a 1x1 filter to reduce the previous layer to 3 dimensions that serves as an interpolated RGB image. 

We utilize the mean-squared-error loss function described in Equation \ref{mse-pixels}. The specifics of our model architecture can be found in Section \ref{sec:algo}, which contains the python code for our best-performing architecture.

\begin{figure}[]
  \centering
  \begin{minipage}[b]{0.4\textwidth}
  	\centering
    \includegraphics[scale=.35]{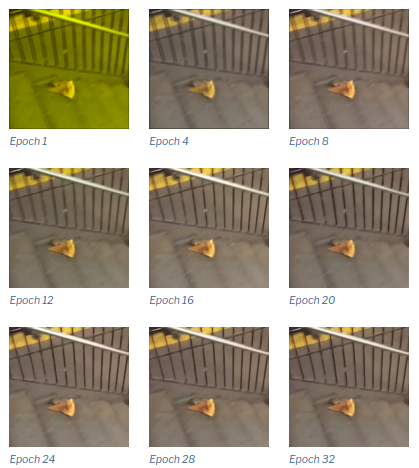}
    \caption{Improvement on 'Pizza Rat' frame interpolation as a function of epochs}
    \label{pizza_rat}
  \end{minipage}
  \hfill
  \begin{minipage}[b]{0.5\textwidth}
  	\centering
    \includegraphics[scale=.6]{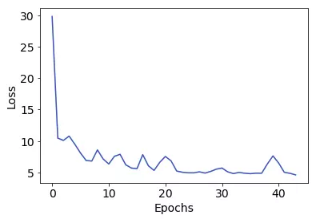}
    \caption{CV Loss Curve}
  \label{cv-loss-curve}
  \end{minipage}
\end{figure}

\section{Results}
\label{results} 

As \cite{Niklaus_encoder}, \cite{gongvideo_autoencoder}, and \cite{nvidia_encoder_decoder} have established, frame interpolation is subject to an extremely qualitative evaluation. We generally found, once a model reaches mean-squared-error value below 7, the loss typically does not reflect qualitative performance. As a result, we will focus on the qualitative results of our top-performing model.

We trained our model on a corpus of diverse videos selected from youtube. Our training data consisted of sets of three frames: the first and the third were used as input to the model, and the third used as ground truth. Our top performing model was trained for 14 hours on a laptop with a graphics card

Figures \ref{pizza_rat} and \ref{cv-loss-curve} visualize the learning curve of our top model performing model. \ref{pizza_rat} visualizes the qualitative improvements of the model, as well as its stagnation, as a function of epochs. \ref{cv-loss-curve} demonstrates the quantitative improvement over epochs as a function of Cross-Validation loss. 

\begin{figure}[]
\centering
    \includegraphics[scale=.7]{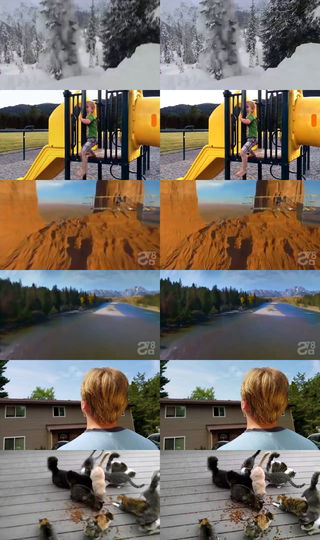}
    
    \centering
    \captionsetup{justification=centering}
    \caption{ \textbf{Left side: Predicted Frame. Right side: Ground Truth Frame} \newline The predict-vs-ground truth frame comparisons enable us to qualitatively evaluate the robustness of our frame interpolation model. The model used to generate these prediction has a mean-squared error of 5.9 between the predicted pixel values and the true pixel values. Visually, there is an issue with slight blurring during rapid motion, but overall, the predicted frames are almost indistinguishable from the ground truth frames.} 
    \label{all_preds}
\end{figure}

In figure \ref{all_preds}, we visually compare our models frame prediction with the validation set ground truth frames. Overall, we obtain almost indistinguishable results. We run into issues with blurring during rapid motion, but we believe that, given our hardware limitations (trained on a laptop), and the relative lack of depth of our model (7 layers), our architecture performs exceptionally well. 

\section{Conclusion}

In this paper, we present a Deep CNN architecture that challenges the current paradigm by which CNN's are designed. In \citeauthor{Kim2014}, word embeddings' lack of impact on CNN NLP models demonstrates that the robustness of CNN's may have nothing to do with their perceived ability to extract local features. \cite{one_pixel_attack} presents an algorithm that consistently alters CNN predictions by changing only 3 pixels in an image. This fragility suggests that pixel importances follow a power-law distribution, similar to word frequencies. As a result, we propose that images ought to be treated not as dense, but rather, effectively sparse data.

In section \ref{nlp_sec}, we discuss work from the field of natural language processing that we used as a conceptual baseline for our model architecture. The most notable trend in NLP models is the consistent and extreme robustness of high-dimensional data representations such as Bag-of-Words combined with simple linear models such as logistic regression. 

In section \ref{model_arch_sec}, we discuss our hypothesis that the robustness of CNNs comes, not from it's ability to detect 'local features,' but rather from a CNN's ability to transform images onto sparse, high dimensional plane. We then present our model architecture that is designed based on this hypothesis.

Figures \ref{pizza_rat} and \ref{cv-loss-curve} visualize our learning curve both quantitatively and qualitatively, and \ref{all_preds} presents our final frame interpolation results. We believe that our results, combined with the relative simplicity of our model demonstrate the validity of our approach to image data and to CNNs. Furthermore, we hope we have effected a certain level of reevaluation concerning the inner workings of convolutional neural networks beyond the phrase 'local feature detection.'

\bibliography{sample.bib}

\section{CNN Model Implementation}
\label{sec:algo}
\begin{lstlisting}[language=Python]
from keras import backend as K
from keras.layers import Conv2D, Dense, Reshape, Lambda, Dropout
from keras.layers.advanced_activations import LeakyReLU
from keras.models import Model, Sequential, load_model, clone_model
from keras.optimizers import Adam

# Load Model
basic = Sequential()

# Convert input to Floating Point type (easier on RAM)
def to_float(x):
    return K.cast(x, "float32")
basic.add(Lambda(to_float, input_shape = (None, None, 6)))

# Add Conv w/in Conv to expand Dims
basic.add(Conv2D(40, (1, 1), 
	strides=(1, 1), padding="SAME", 
	input_shape=(None, None, 6)))
basic.add(LeakyReLU())

# Dimensionality Expanders
basic.add(Conv2D(40, (7, 7), 
			strides=(1, 1), activation=None, padding="SAME"))
basic.add(LeakyReLU())
basic.add(Conv2D(64, (4, 4), 
	strides=(1, 1), activation=None, padding="SAME"))
basic.add(LeakyReLU())
basic.add(Conv2D(70, (3, 3), 
	strides=(1, 1), activation=None, padding="SAME"))
basic.add(LeakyReLU())
basic.add(Conv2D(70, (2, 2), 
	strides=(1, 1), activation=None, padding="SAME"))
basic.add(LeakyReLU())

# Dense NN to Reduce Dimensionality
basic.add(Conv2D(80, (1, 1), 
	strides=(1, 1), activation=None, padding="SAME"))
basic.add(LeakyReLU())
basic.add(Conv2D(28, (1, 1), 
	strides=(1, 1), activation=None, padding="SAME"))
basic.add(LeakyReLU())
basic.add(Conv2D(3, (1, 1), 
	strides=(1, 1), activation='relu', padding="SAME"))
\end{lstlisting}







\end{document}